\documentclass[conference]{IEEEtran}
\IEEEoverridecommandlockouts
\usepackage{cite}
\usepackage{amsmath,amssymb,amsfonts}
\usepackage{algorithmic}
\usepackage{graphicx}
\usepackage{textcomp}
\usepackage{xcolor}
\usepackage{subfig}  
\usepackage{booktabs}  
\usepackage{float}  

\makeatletter
\let\NAT@parse\undefined
\makeatother
\usepackage{hyperref}  
\hypersetup{hidelinks}

\def\BibTeX{{\rm B\kern-.05em{\sc i\kern-.025em b}\kern-.08em
    T\kern-.1667em\lower.7ex\hbox{E}\kern-.125emX}}
\begin{document}

\title{Real Time Bearing Fault Diagnosis Based on Convolutional Neural Network and STM32 Microcontroller}

\author{\IEEEauthorblockN{Wenhao Liao}
\IEEEauthorblockA{\textit{Shenzhen Institute for Advanced Study} \\
\textit{University of Electronic Science and Technology of China}\\
Shenzhen 518110, China \\whliao@std.uestc.edu.cn
}}


\maketitle

\begin{abstract}
With the rapid development of big data and edge computing, many researchers focus on improving the accuracy of bearing fault classification using deep learning models, and implementing the deep learning classification model on limited resource platforms such as STM32. To this end, this paper realizes the identification of bearing fault vibration signal based on convolutional neural network, the fault identification accuracy of the optimised model can reach 98.9\%. In addition, this paper successfully applies the convolutional neural network model to STM32H743VI microcontroller, the running time of each diagnosis is 19ms. Finally, a complete real-time communication framework between the host computer and the STM32 is designed, which can perfectly complete the data transmission through the serial port and display the diagnosis results on the TFT-LCD screen.
\end{abstract}


\section{Introduction}
For the mechanical equipment that can be seen everywhere now, its development has become more and more complex and systematic trend, and a huge mechanical system, any small part failure may lead to the whole system equipment as a whole collapse, resulting in serious economic losses and huge casualties, so monitoring the health status of the mechanical operation process is of great significance for cost control and production safety. Bearings in large equipment are an extremely important class of parts, known as the joints of modern machinery, whose main function is to support the mechanical rotating body and reduce the coefficient of friction in the process of movement. 

Traditional fault diagnosis methods are based on empirical methods, and the collected data is processed and displayed in various ways, and the fault is determined by the manual experience of the engineer, which is becoming increasingly difficult in the face of increasingly complex and sophisticated mechanical equipment. Deep learning is a data-driven approach that only requires accurate raw data and a suitable network structure to achieve end-to-end automatic fault diagnosis. With the advent of the big data era, there are now low-cost and accurate sensors that can collect and monitor data on the operation of bearings, while data-driven deep learning algorithms are being used for fault detection due to their powerful feature extraction capabilities, the results are more accurate and less expensive per unit than traditional methods.

Based on the above discussion, this paper proposes that real time bearing fault diagnosis based on convolutional neural network and STM32 microcontroller, This study has the following main contributions:

\begin{enumerate}
    \item In this study, a convolutional neural network is used to achieve a state-of-the-art accuracy of up to 98.9\% for identifying 10 different types of bearing failure in the CWRU dataset.
    \item The lightweight convolutional neural network model proposed in this study has a 3.8\% improvement in accuracy and a 36.66\% reduction in the total number of network parameters compared to conventional convolutional neural network models, making it more suitable for running in low-power devices such as the STM32.
    \item This study successfully identifies bearing faults by embedding convolutional neural networks into the STM32 using the CubeAI tool.
    \item This study implements a complete convolutional neural network and STM32 based real-time bearing fault detection process framework, and in a subsequent validation session, demonstrates that it can perfectly achieve the bearing detection objectives.
\end{enumerate}

\section{Related Work}
Lei et al.\cite{leiApplicationsMachineLearning2020}, by collecting the number of published papers, divided the development of this field of intelligent fault diagnosis into roughly three phases: the first phase is the traditional machine learning-based diagnosis methods starting from 1980, including support vector machine (SVM), K-nearest neighbour algorithm (KNN), probabilistic graphical model (PGM), random forest algorithm (RF), etc.; the second phase is the deep learning-based diagnosis methods starting from 2010. The second phase is the deep learning based diagnostic methods that started in 2010, including deep neural networks (DNN), deep confidence networks (DBN), convolutional neural networks (CNN), and recurrent neural networks (RNN), etc. The third phase is the migration learning methods that started in 2010, including deep learning based diagnostic methods that started in 2010, including deep neural networks (DNN), deep confidence networks (DBN), convolutional neural networks (CNN), and recurrent neural networks (RNN), etc. The third phase is the migration learning methods that have emerged since 2015, including adversarial neural networks (GAN) and joint adaptive networks (JAN), etc. This year, with the rise of reinforcement learning and migration learning, there are three major trends in the future development of intelligent fault diagnosis: first, most practical engineering applications are in high-noise environments, and how to denoise to improve model recognition accuracy. Finally, there is the research on breakthrough theoretical algorithms, including new automatic noise reduction network structures, signal feature detection structures, etc, such as the one proposed by Zhao Minghang et al.\cite{zhaoDeepResidualShrinkage2020} in China for high noise environments. proposed a deep residual systolic network (including both DRSN-CS and DRSN-CW structures) for high accuracy fault diagnosis in high noise environments.

\section{Model Design}
Our aim is to design a complete process framework for bearing fault detection, implement a highly accurate convolutional neural network model deployed on the STM32, and use serial communication between the host computer and the STM32 to transmit bearing vibration data for real-time bearing fault detection functions.

the overall model design is illustrated in Fig.~\ref{fig:figure1}, which is divided into five modules. Among them, the data processing module mainly extracts the bearing fault data from the CWRU dataset in mat format and then performs normalised pre-processing; the training module obtains the trained model by constructing a convolutional neural network model and using the processed data for training; the serial communication module uses the designed host computer to transmit the bearing fault data to the STM32 via the serial port in real time and uses a GUI to display the process of data transmission; the result display module uses the results obtained by the prediction module to display the prediction results and the prediction consumption time on a TFT-LCD screen.

\subsection{Data Processing Module}
\begin{figure}[h]
	\centering
	\subfloat[Bearing test rig of CWRU]{\includegraphics[width=0.4\textwidth]{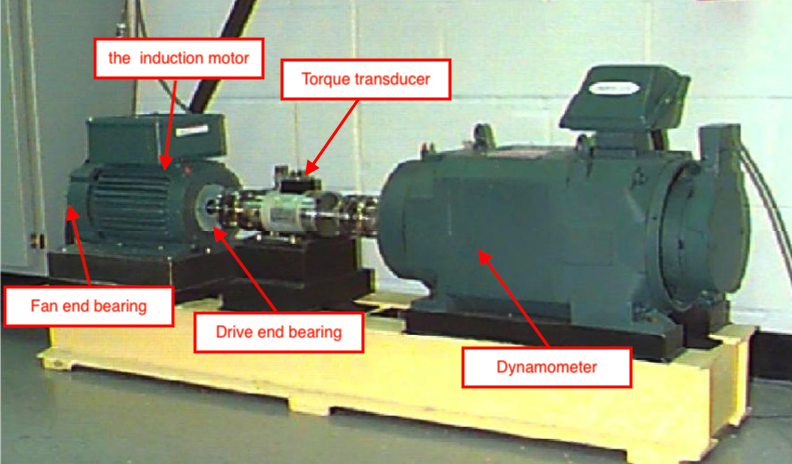}}
    \hspace{0.8in}
	\subfloat[Components of bearing. Referenced from \cite{boudiafComparativeStudyVarious2016}]{\includegraphics[width=0.4\textwidth]{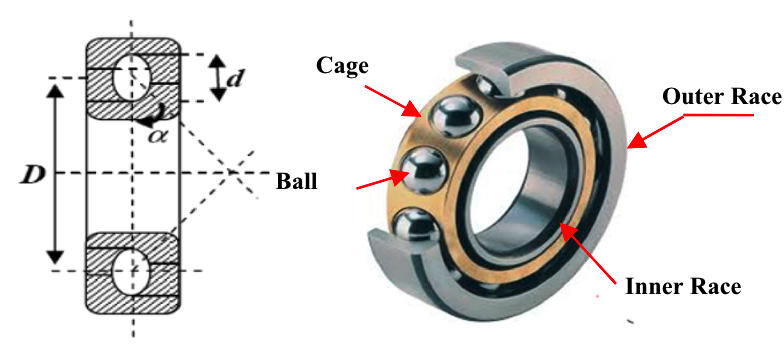}}
	\caption{Experimental platform of CWRU}
	\label{fig:figure2}
\end{figure}

\begin{figure*}[thbp]
    \centering
    \includegraphics[width=\textwidth,height=0.4\textwidth]{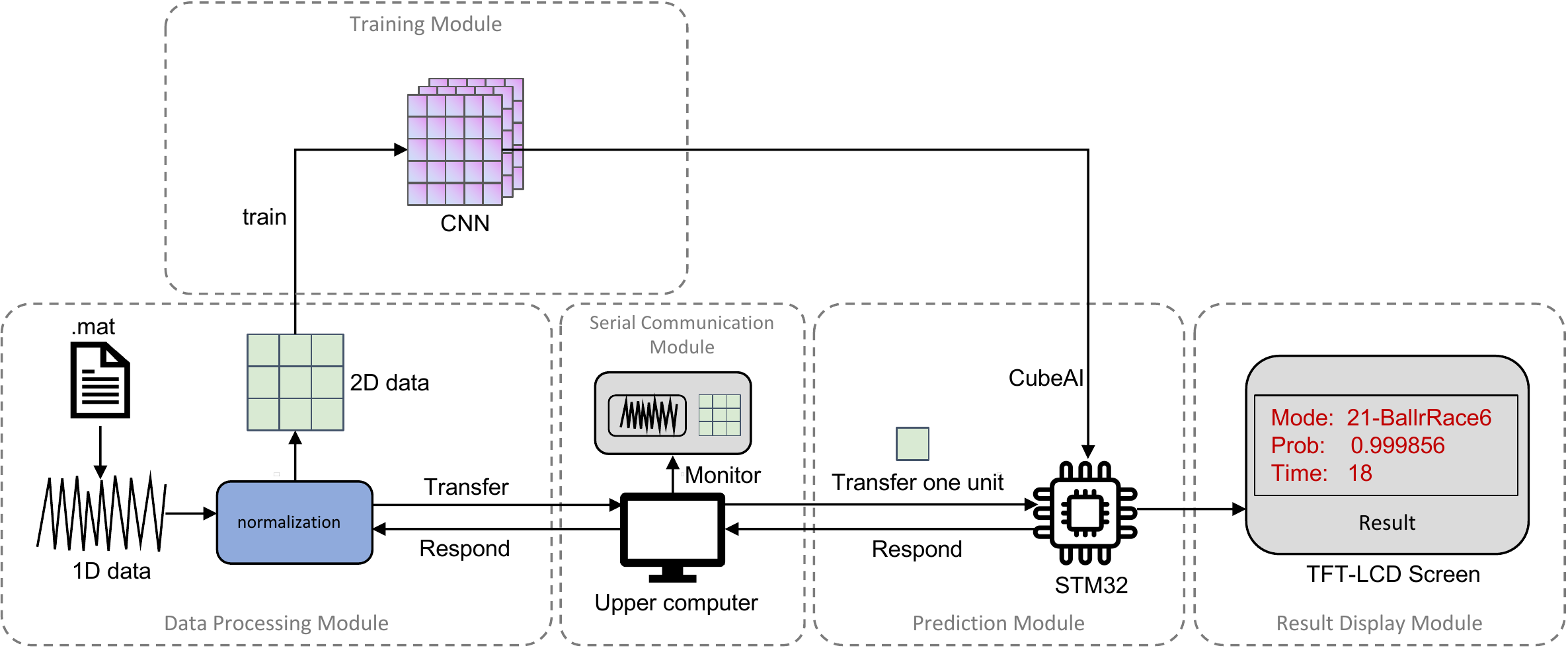}
    \caption{A overall architecture of the proposed model.}\label{fig:figure1}
\end{figure*}

The CWRU Bearing Dataset\cite{BearingDataCenter2021} is a dataset of failed bearings published by the Case Western Reserve University Data Centre in the USA. Its purpose is to test and verify engine performance and in recent years, due to the growth of fault diagnosis, it has been used mainly as benchmark data for fault signal diagnosis. The experimental platform for the measurement data and the rolling bearing structure used in the experiments are shown in Fig.~\ref{fig:figure2}.

For the entire CWRU faulty bearing data set, the drive side data at the 12 KHz sampling rate is the most complete and has the least obvious human error. For the data at different speeds, the data at 1797 rpm is closest to 1800 rpm, at which point the data at 400 sampling points is close enough to a full revolution for a complete fault signal to be observed in the signal at the 12 KHz sampling rate\cite{smithRollingElementBearing2015}. Therefore, the subsequent training and validation of the network structure is carried out using the drive end data of 1797 rpm at the 12 KHz sampling rate to obtain 10 different fault classifications as shown in Table~\ref{tab:table1}.

\begin{table}[htbp]
    \centering
    \caption{10 different fault classifications obtained from the CWRU dataset.}
    \label{tab:table1}
    \begin{tabular}{c l l}
        \toprule
        Number & Parameter name & Annotation \\ \midrule
        1 & 00-Normal & Normal without fault \\
        2 & 07-Ball & 0.007 inch ball fault \\
        3 & 07-InnerRace & 0.007 inch inner race fault \\
        4 & 07-OuterRace6 & 0.007 inch 6 o'clock race fault \\
        5 & 14-Ball & 0.014 inch ball fault \\
        6 & 14-InnerRace & 0.014 inch inner race fault \\
        7 & 14-OuterRace6 & 0.014 inch 6 o'clock race fault  \\
        8 & 21-Ball & 0.021 inch ball fault \\
        9 & 21-InnerRace & 0.021 inch inner race fault \\ 
        10 & 21-OuterRace6 & 0.021 inch 6 o'clock race fault \\ 
        \bottomrule
    \end{tabular}
\end{table}

\begin{figure}[h]
	\centering
    \includegraphics[width=0.4\textwidth]{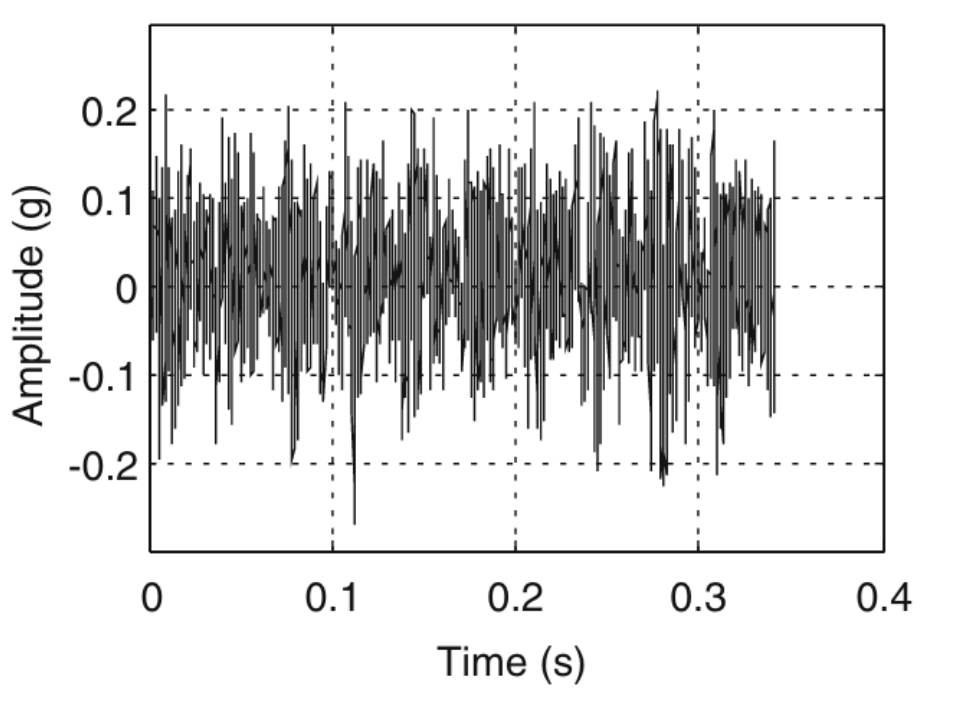}
	\caption{Part of the cwru dataset data under time series.}
	\label{fig:figure3}
\end{figure}

The required bearing vibration signal data is then extracted as described above for subsequent analysis and processing. Fig.~ref{fig:figure3} illustrates the transformation of a section of signal data into a time series. Further pre-processing of the signal data, such as data denoising, data sampling and data normalisation, is required according to the needs of the experiment.

For the denoising problem, manual denoising is not required because the CWRU dataset is a manual EDM pointing simulation of the defect, which has a low noise level\cite{smithRollingElementBearing2015}. In this paper, we use random sampling to eliminate as much as possible the case of data preference of the training data. In order to optimise the data distribution for different fault cases and to improve the accuracy and training speed of the network model, the data features are subjected to dimensionless processing, i.e. normalisation, and the normalisation method used here is the min-max method, the formula is obtained as the follows:

\begin{equation}
    y=\dfrac{x-min\left(x\right)}{max\left(x\right)-min\left(x\right)}
\end{equation}

Bearing fault signals are one-dimensional time series of data that need to be transformed into a two-dimensional matrix for training with CNN networks.

\begin{figure}[h]
	\centering
    \includegraphics[width=0.4\textwidth]{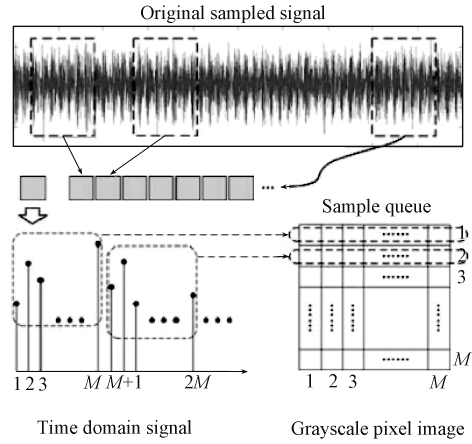}
	\caption{The conversion of one dimension signal to image. Referenced from \cite{refwuJiyujuanjishenjingwangluodejixieguzhangzhenduanfangfazongshu2020}.}
	\label{fig:figure4}
\end{figure}

The transformation method we use here is the direct transformation method, which directly intercepts a certain length of the acquired 1D data through a sliding window, and then stacks the intercepted data multiple times to obtain a 2D matrix data. The simplicity of this method makes it suitable for use in real-time fault detection systems for low-power devices, as shown in Fig.~\ref{fig:figure4}.

\subsection{Training Module}
This module trains the CNN model with the 2D matrix data obtained in the data processing module, we selected Tanh as the network activation function and added the ReduceLR adaptive learning rate adjustment mechanism. After adjusting the parameters and network structure for several times and training, we obtained a satisfactory network model with the final accuracy, whose network structure and parameters are shown in Table~\ref{tab:table2}.

\begin{table}[htbp]
    \centering
    \caption{Our CNN model parameter details.}
    \label{tab:table2}
    \begin{tabular}{cccc}
        \toprule
        Layer name & Filter size & Filter number & Stride  \\
        \midrule
        Conv1 & (10, 10) & 4 & 1  \\ 
        Conv2 & (5, 5) & 8 & 1  \\ 
        Maxpool1 & (4, 4) & 8 & 2  \\ 
        Conv3 & (3, 3) & 16 & 1  \\ 
        Conv4 & (3, 3) & 16 & 1  \\ 
        Maxpool2 & (2, 2) & 16 & 2  \\ 
        Conv5 & (3, 3) & 32 & 1  \\ 
        Conv6 & (3, 3) & 64 & 1  \\ 
        Maxpool3 & (1, 1) & 64 & 2  \\ 
        Full connection & 32 & - & -  \\ 
        Softmax & 10 & - & - \\
        \bottomrule
    \end{tabular}
\end{table}

We can then apply the trained model to the bearing fault detection task and use it in a subsequent prediction module on the STM32.

\subsection{Prediction Module}
In this module we use the CubeAI tool to deploy the trained models from the training module to the STM32. CubeAI is an official ST AI extension package for CubeMX that supports 8-bit quantization of Keras networks and TensorFlowLite quantization networks by automatically transforming pre-trained neural networks and integrating the resulting optimisation libraries into the user's project\cite{XCUBEAIAIExpansion}.

When importing the model, there is an option to compress and quantise the weights and parameters, but this will reduce the accuracy of the model. The STM32H743VI microcontroller chosen for this work has sufficient resources, so the network model is not compressed and quantized. After importing the model, the complexity of the network parameters is 1238380 MACC, MACC (multiply-accumulate operations) is used to describe the model complexity of the neural network, which represents the number of operations that are multiplied and then added in the network calculation. The specific resource consumption is shown in Table~\ref{tab:table3}.

\begin{table}[htbp]
    \centering
    \caption{A simple table with a header row.}
    \label{tab:table3}
    \begin{tabular}{c c}
        \toprule
        Index & Value \\ \midrule
        MACC & 1238380 \\
        FLASH & 142.15KB \\
        RAM & 11.32KB \\
        \bottomrule
    \end{tabular}
\end{table}

\begin{figure*}[thbp]
	\centering
    \includegraphics[width=\textwidth]{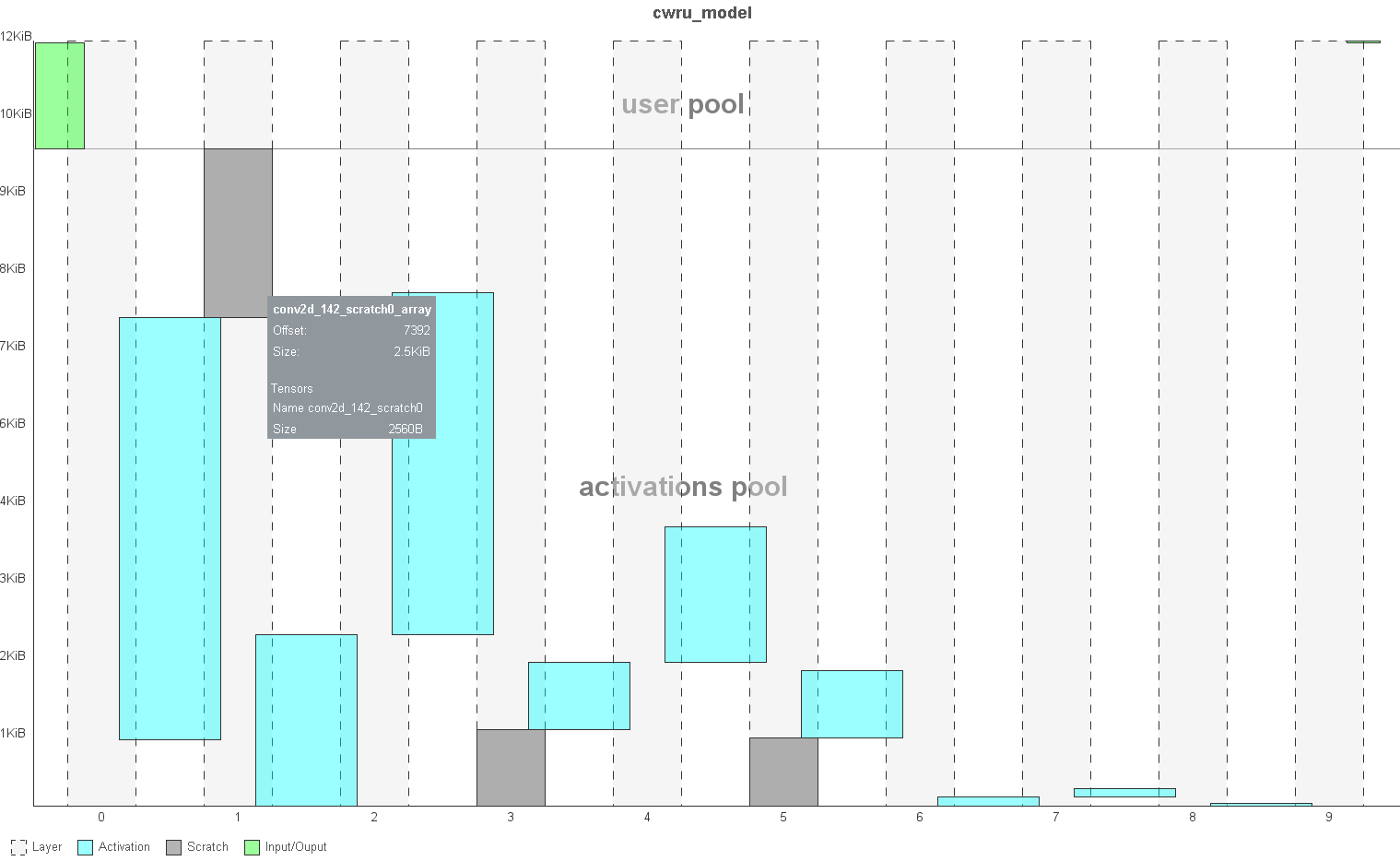}
	\caption{The specific RAM consumption per layer.}
	\label{fig:figure5}
\end{figure*}

Although the initialisation and definition of the subsequent input and intermediate variables require system resources, they are not significant compared to the 2M FLASH and 1M RAM of the STM32H743VI microcontroller. The specific RAM consumption per layer is shown in Fig.~\ref{fig:figure5}. As we can see from Fig.~\ref{fig:figure5}, the activation function in the network model consumes most of the RAM, and the RAM consumption of all the layers together must be more than 12KB, but since the network model can be deleted from RAM after one layer and then the next layer can be run, we only need to calculate the maximum RAM consumption value during one layer to get the maximum RAM consumption value.

\subsection{Serial Communication Module}
The main purpose of this module is to implement the transfer of bearing fault data via serial communication to the STM32 microcontroller to facilitate subsequent fault detection tasks.

\begin{figure*}[thbp]
	\centering
    \includegraphics[width=\textwidth]{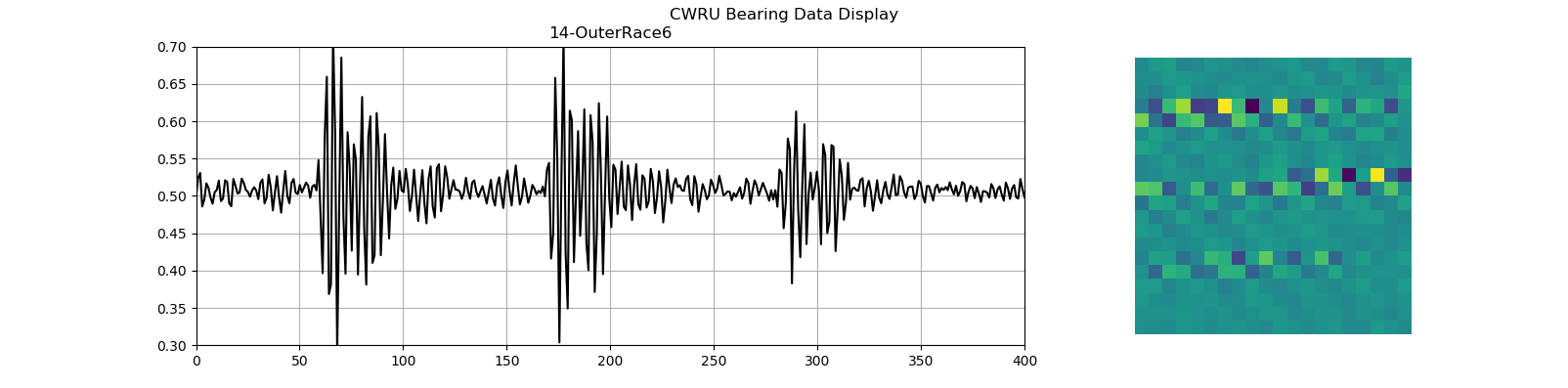}
	\caption{Real-time data display interface of host computer.}
	\label{fig:figure6}
\end{figure*}

Here we have developed a higher level computer software to facilitate the serial communication between the PC and the microcontroller. As the data sent by serial communication is ASCII and the data transferred is in character format rather than floating point format, the transfer process requires conversion and recovery of floating point numbers. To facilitate the visualisation of the data, we have also designed a display interface for the real-time data, as shown in Fig.~\ref{fig:figure6}, On the left is the one dimensional time series data and on the right is the two dimensional data obtained by converting the one dimensional data using the direct conversion method.

\begin{figure}[H]
	\centering
    \includegraphics[width=0.4\textwidth]{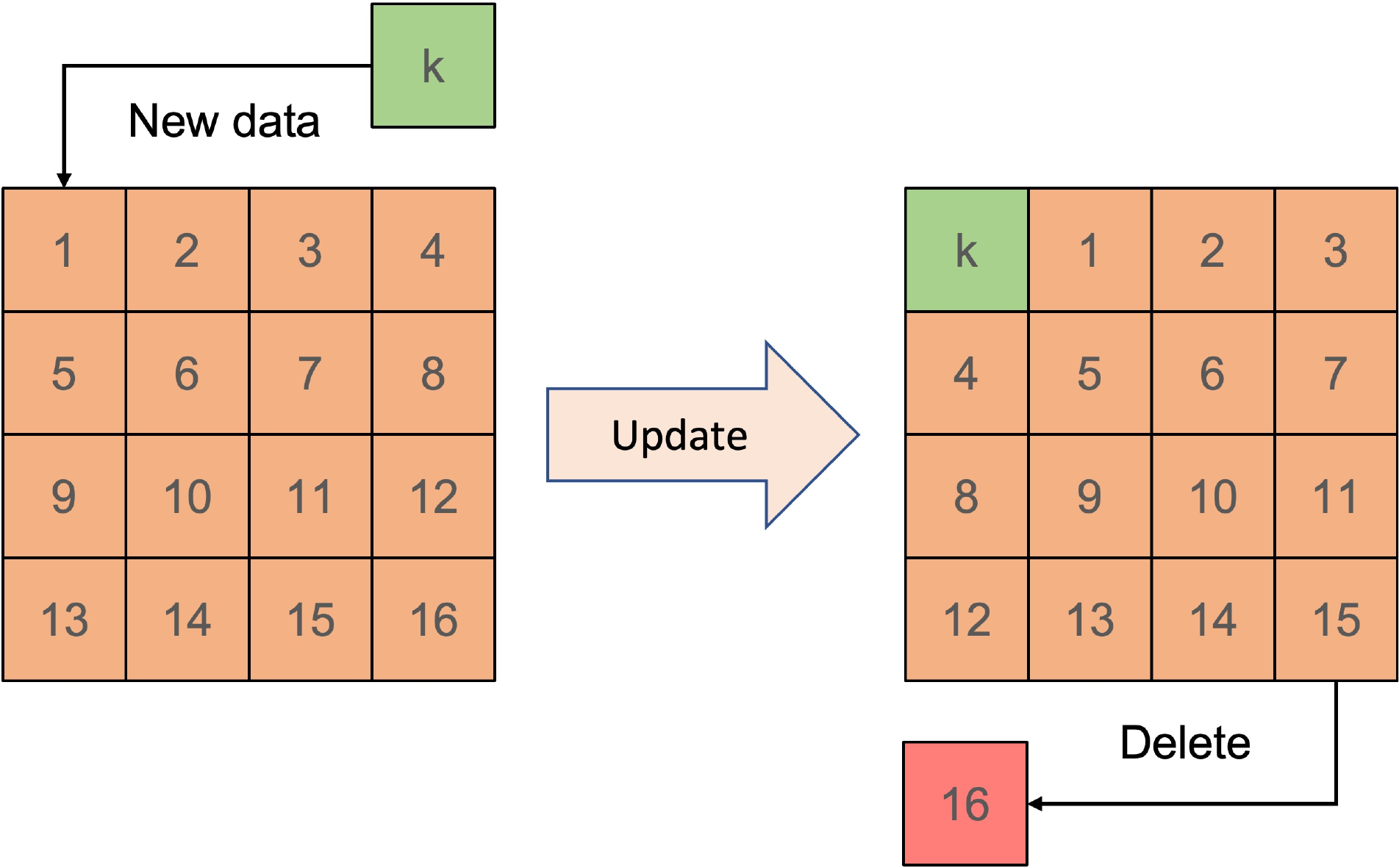}
	\caption{The update process of buffer in STM32.}
	\label{fig:figure7}
\end{figure}

A 20 * 20 buffer is set up on the STM32 microcontroller to hold the data. The update process after receiving new data from the serial port is shown in Fig.~\ref{fig:figure7}, for demonstration purposes only a 4 * 4 matrix is used. 

The data in the upper left corner of the matrix is the input port and the data in the lower right corner is the output port. After receiving the new data, the data in the matrix is first moved from left to right, with the rightmost data in each row moving to the leftmost data in the next row and the rightmost data in the last row being discarded.

The prediction module then performs a new round of predictions on the data in the buffer. The prediction results from the completion of the network model are transferred to the display module for display and then returned to the host computer via the serial communications module to indicate that the prediction has been completed and the next data transfer can begin.

\subsection{Result Display module}
\begin{figure}[t]
	\centering
    \includegraphics[width=0.3\textwidth]{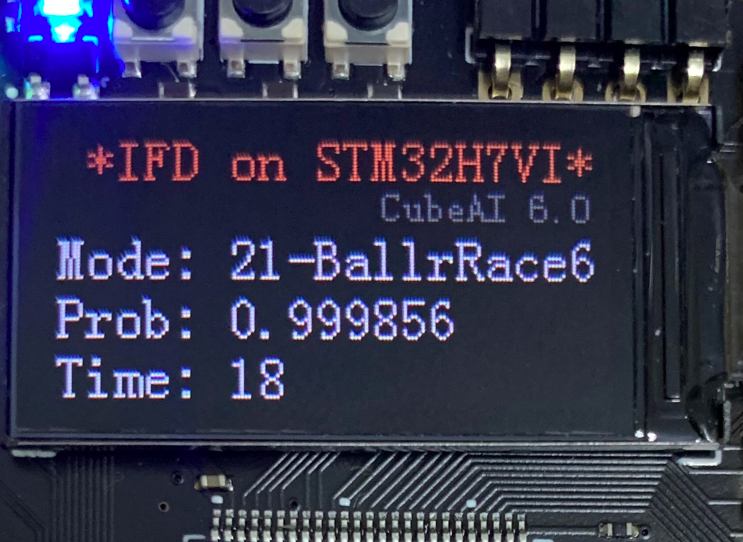}
	\caption{A example display of a prediction.}
	\label{fig:figure8}
\end{figure}

This module displays on a TFT-LCD screen the predicted failure type, failure probability and predicted process time obtained by the prediction module. The example display is shown in Fig.~\ref{fig:figure8}.

The contents of the figure indicate that the diagnostic fault result is a 21 inch fault on the bearing ball and that the probability of this fault is 0.999856 and the time taken for this diagnostic result is 18 ticks.

\section{Experimental Test}
The chip model used in this experiment is the STM32H743VI, a high performance ARM Cortex-M7 MCU with DSP and DP-FPU with 2 MB Flash, 1 MB RAM, 480 MHz CPU. Use CubeMX to configure the chip, CubeIDE to write the relevant control programmes, and the CubeAI plug-in to deploy convolutional neural network models in the STM32.

We conducted comparative experiments on the selection of CNN models, where the benchmark model is the CNN model in the official Keras MNIST handwriting recognition example; the Tanh model is based on the benchmark model by changing the ReLU activation function to the Tanh activation function; and the ReduceLR model is based on the benchmark model by adding an adaptive learning rate adjustment mechanism.

The Tanh activation function can maintain a non-linear monotonic ascending and descending relationship between output and input, which is consistent with the gradient solution of BP (back-propagation) networks, is fault-tolerant and bounded, but takes longer to compute the derivative than the ReLU function. The adaptive learning rate adjustment mechanism is to have a larger learning rate at the beginning of training and reduce the learning rate as the metrics improve, to have a larger learning rate at the beginning of training so that the network model can be fitted faster, and to reduce the learning rate as the network metrics continue to increase, to avoid network oscillation and delayed fitting.

\begin{figure}[t]
	\centering
	\subfloat[Accuracy of three comparison models during training]{\includegraphics[width=0.4\textwidth]{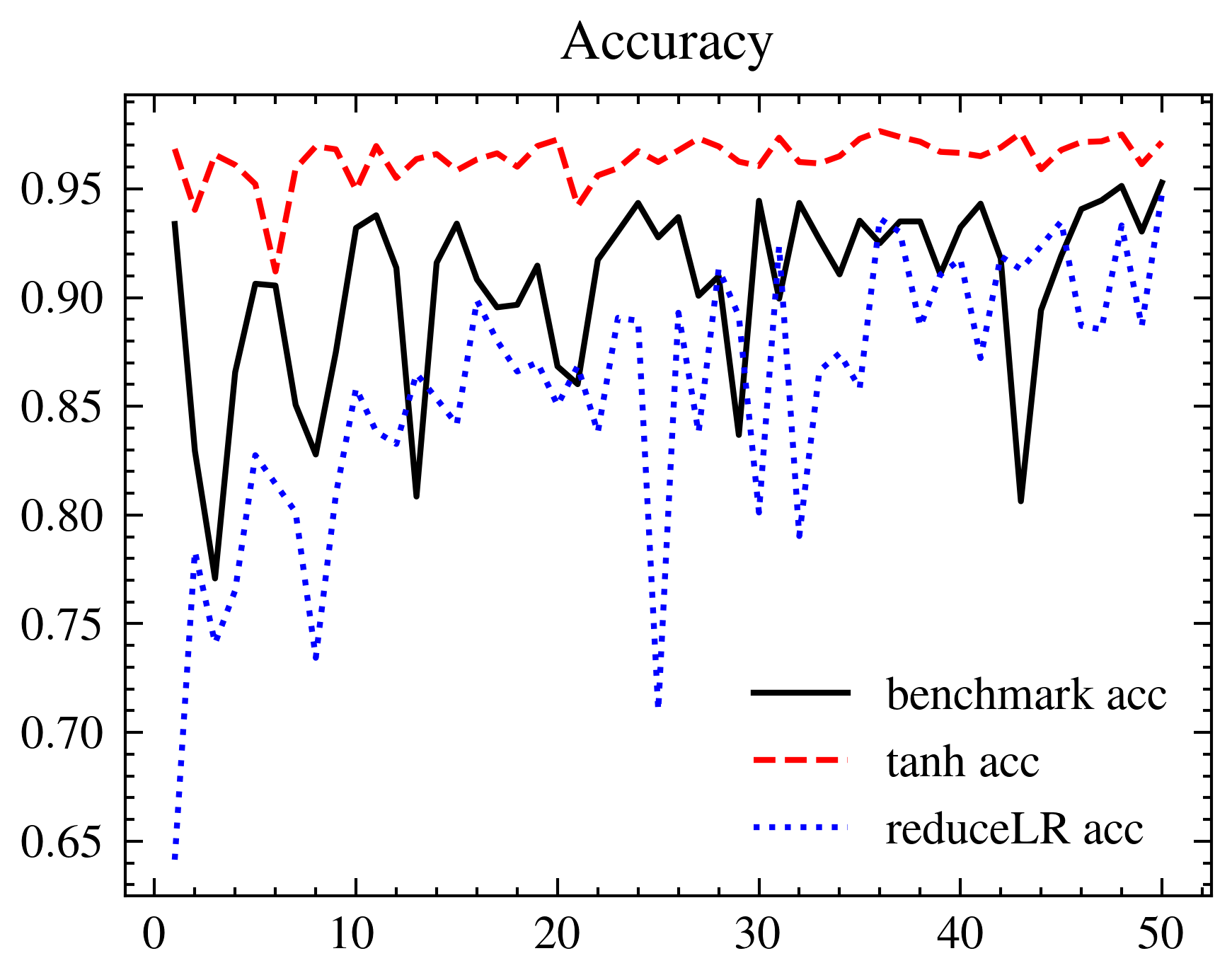}}
    \hspace{0.8in}
	\subfloat[Accuracy of our optimised model during training]{\includegraphics[width=0.4\textwidth]{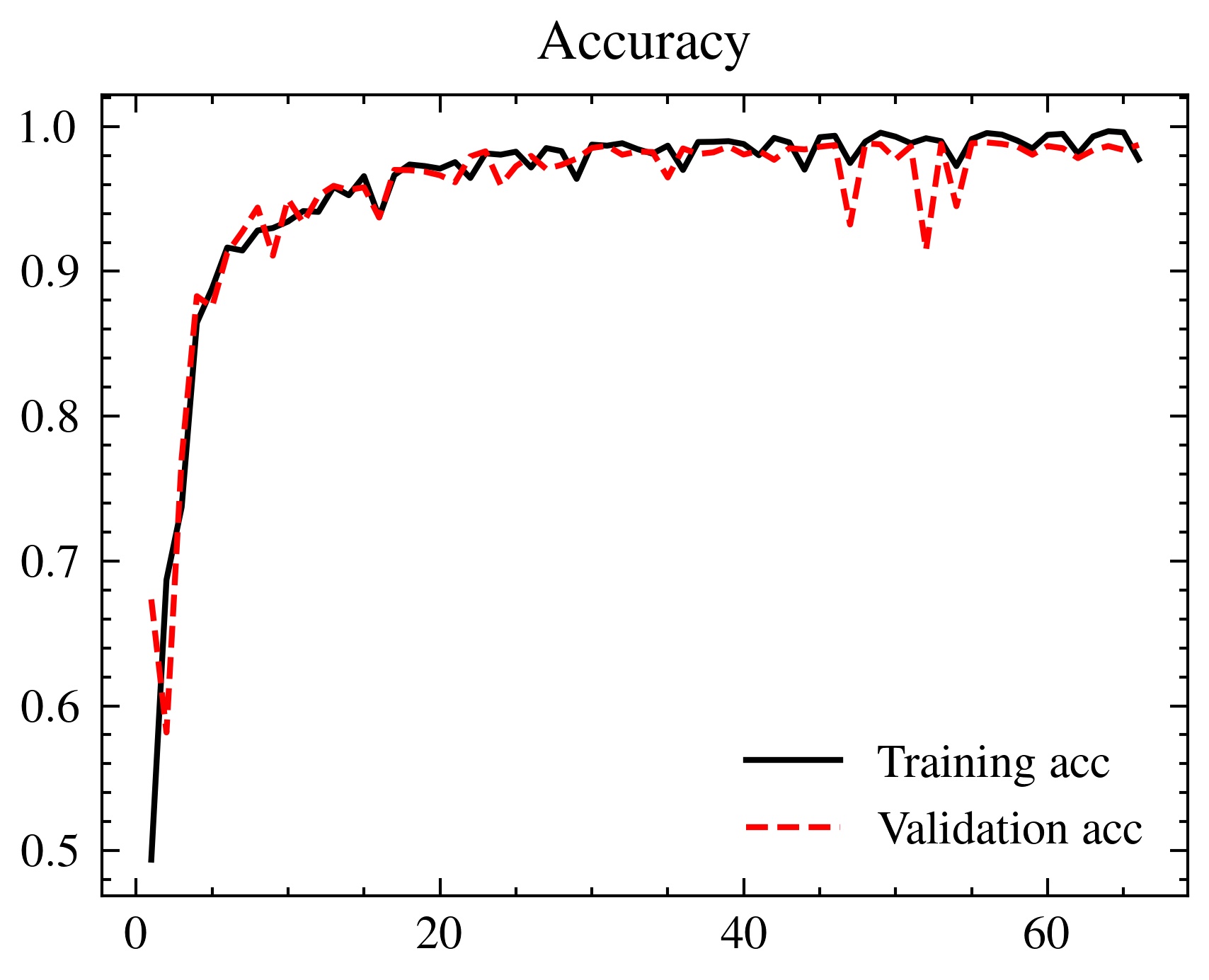}}
	\caption{Accuracy of models during training.}
	\label{fig:figure9}
\end{figure}

The accuracy of the three models compared and the optimised model proposed in this paper are shown in Fig.~\ref{fig:figure9}, from which (a) we can see that the benchmark solution has very strong oscillations in the training process, while the solution using Tanh as the activation function is the most stable and accurate. The overall trend is a gradual fit and the oscillations are gradually reduced.

Therefore, our proposed optimisation network uses Tanh as the activation function and adds the ReduceLR adaptive learning rate adjustment mechanism to optimise the fitting trend of the network. As can be seen in (b), the validation accuracy during training of our model follow the training accuracy and loss well, which means that the network model is neither overfitting nor underfitting, and the accuracy is higher than the three previous comparison models.

Some performance comparisons between our optimised model and the three previous comparison models are shown in Table~\ref{tab:table4}. It can be seen that our optimised model has the lowest total number of parameters and the best performance among them in terms of training time, best accuracy and minimum loss value, indicating that our model can save the most resources to achieve optimal results.

\begin{table}[!ht]
    \centering
    \caption{CNN model parameter details}
    \label{tab:table4}
    \scalebox{0.86}{  
    \begin{tabular}{cccc}
        \toprule
        Model & Total parameters & Training time / round & best accuracy \\
        \midrule
        Benchmark model & 55,690 & 2.2s & 0.9528 \\ 
        Tanh model & 55,690 & 2.2s & 0.9765 \\ 
        ReduceLR model & 55,690 & 2.3s & 0.9477 \\ 
        \textbf{our optimised model} & \textbf{36,390} & \textbf{2.0s} & \textbf{0.9890} \\
        \bottomrule
    \end{tabular}
    }
\end{table}

\section{Conlusion}
In this paper, we experimentally obtain a lightweight CNN model that can be used for bearing fault detection with high accuracy, and successfully deploy it on a terminal device such as STM32. A convolutional neural network and STM32-based real-time bearing fault detection process framework, including data pre-processing and transmission, model training and deployment, real-time data monitoring and fault detection, and fault detection result presentation, has also been implemented and validated for usability.

\bibliographystyle{IEEEtran}
\bibliography{ref,FaultDiagnosis}

\end{document}